\documentclass{bmcart}

\usepackage[utf8]{inputenc} %unicode support
\usepackage{graphicx}
\usepackage{amsmath}
\usepackage{amsfonts}
\usepackage{subfigure}
\usepackage{caption}
\usepackage{indentfirst}
\usepackage{cite}
\usepackage[caption=false]{subfig}

\startlocaldefs
\endlocaldefs

\begin{document}

%%% Start of article front matter
\begin{frontmatter}

\begin{fmbox}
\dochead{Research}


\title{Hybrid Attentional Memory Network for Computational drug repositioning}

%%%%%%%%%%%%%%%%%%%%%%%%%%%%%%%%%%%%%%%%%%%%%%
%%                                          %%
%% Enter the authors here                   %%
%%                                          %%
%% Specify information, if available,       %%
%% in the form:                             %%
%%   <key>={<id1>,<id2>}                    %%
%%   <key>=                                 %%
%% Comment or delete the keys which are     %%
%% not used. Repeat \author command as much %%
%% as required.                             %%
%%                                          %%
%%%%%%%%%%%%%%%%%%%%%%%%%%%%%%%%%%%%%%%%%%%%%%
\author[
addressref={1},
email={jieyuehe@seu.edu.cn},
noteref={n1},
]{\inits{HJY}\fnm{Jieyue} \snm{He}}
\author[
addressref={1},
noteref={n1},
email={220174323@seu.edu.cn}
]{\inits{YXX}\fnm{Xinxing} \snm{Yang}}
\author[
addressref={1},
email={}
]{\inits{GZ}\fnm{Zhuo} \snm{Gong}}
\author[
addressref={1},                   % id's of addresses, e.g. {aff1,aff2}
corref={1},                       % id of corresponding address, if any
%noteref={n1},                        % id's of article notes, if any
email={jieyuehe@seu.edu.cn}   % email address
]{\inits{ZI}\fnm{lbrahim} \snm{Zamit}}

%%%%%%%%%%%%%%%%%%%%%%%%%%%%%%%%%%%%%%%%%%%%%%
%%                                          %%
%% Enter the authors' addresses here        %%
%%                                          %%
%% Repeat \address commands as much as      %%
%% required.                                %%
%%                                          %%
%%%%%%%%%%%%%%%%%%%%%%%%%%%%%%%%%%%%%%%%%%%%%%

\address[id=1]{%                           % unique id
	\orgname{School of Computer Science and Engineering, Key Lab of Computer
		Network \& Information Integration,  MOE, Southeast University}, % university, etc                    %
	\postcode{210018},                              % post or zip code
	\city{Nanjing},                              % city
	\cny{China}                                    % country
}

%%%%%%%%%%%%%%%%%%%%%%%%%%%%%%%%%%%%%%%%%%%%%%
%%                                          %%
%% Enter short notes here                   %%
%%                                          %%
%% Short notes will be after addresses      %%
%% on first page.                           %%
%%                                          %%
%%%%%%%%%%%%%%%%%%%%%%%%%%%%%%%%%%%%%%%%%%%%%%

\begin{artnotes}
%\note{Sample of title note}     % note to the article
\note[id=n1]{Equal contributor. These authors contributed equally to this study and share first authroship} % note, connected to author
\end{artnotes}

\end{fmbox}% comment this for two column layout

%%%%%%%%%%%%%%%%%%%%%%%%%%%%%%%%%%%%%%%%%%%%%%
%%                                          %%
%% The Abstract begins here                 %%
%%                                          %%
%% Please refer to the Instructions for     %%
%% authors on http://www.biomedcentral.com  %%
%% and include the section headings         %%
%% accordingly for your article type.       %%
%%                                          %%
%%%%%%%%%%%%%%%%%%%%%%%%%%%%%%%%%%%%%%%%%%%%%%

\begin{abstractbox}

\begin{abstract} % abstract
\parttitle{Background} %if any
Drug repositioning has been an important and efficient method for discovering new uses of known drugs. Researchers have been limited to one certain type of Collaborative Filtering (CF) models for drug repositioning, like the neighborhood based approaches which are good at mining the local information contained in few strong drug-disease associations, or the latent factor based models which are effectively capture the global information shared by a majority of drug-disease associations. Few researchers have combined these two types of CF models to derive a hybrid model which can offer the advantages of both. Besides, the cold start problem has always been a major challenge in the field of computational drug repositioning, which restricts the inference ability of relevant models.	

\parttitle{Results} %if any
Inspired by the memory network, we propose the Hybrid Attentional Memory Network (HAMN) model, a deep architecture combining two classes of CF models in a nonlinear manner. First, the memory unit and the attention mechanism are combined to generate a neighborhood contribution representation to capture the local structure of few strong drug-disease associations. Then a variant version of the autoencoder is used to extract the latent factor of drugs and diseases to capture the overall information shared by a majority of drug-disease associations. During this process, ancillary information of drugs and diseases can help alleviate the cold start problem. Finally, in the prediction stage, the neighborhood contribution representation is coupled with the drug latent factor and disease latent factor to produce predicted values. Comprehensive experimental results on two data sets demonstrate that our proposed HAMN model outperforms other comparison models based on the AUC, AUPR and HR indicators.
\parttitle{Conclusions} %if any
Through the performance on two drug repositioning data sets, we believe that the HAMN model proposes a new solution to improve the prediction accuracy of drug-disease associations and give pharmaceutical personnel a new perspective to develop new drugs.
\end{abstract}

%%%%%%%%%%%%%%%%%%%%%%%%%%%%%%%%%%%%%%%%%%%%%%
%%                                          %%
%% The keywords begin here                  %%
%%                                          %%
%% Put each keyword in separate \kwd{}.     %%
%%                                          %%
%%%%%%%%%%%%%%%%%%%%%%%%%%%%%%%%%%%%%%%%%%%%%%


% MSC classifications codes, if any
%\begin{keyword}[class=AMS]
%\kwd[Primary ]{}
%\kwd{}
%\kwd[; secondary ]{}
%\end{keyword}

\end{abstractbox}
%
%\end{fmbox}% uncomment this for twcolumn layout

\end{frontmatter}

%%%%%%%%%%%%%%%%%%%%%%%%%%%%%%%%%%%%%%%%%%%%%%
%%                                          %%
%% The Main Body begins here                %%
%%                                          %%
%% Please refer to the instructions for     %%
%% authors on:                              %%
%% http://www.biomedcentral.com/info/authors%%
%% and include the section headings         %%
%% accordingly for your article type.       %%
%%                                          %%
%% See the Results and Discussion section   %%
%% for details on how to create sub-sections%%
%%                                          %%
%% use \cite{...} to cite references        %%
%%  \cite{koon} and                         %%
%%  \cite{oreg,khar,zvai,xjon,schn,pond}    %%
%%  \nocite{smith,marg,hunn,advi,koha,mouse}%%
%%                                          %%
%%%%%%%%%%%%%%%%%%%%%%%%%%%%%%%%%%%%%%%%%%%%%%

%%%%%%%%%%%%%%%%%%%%%%%%% start of article main body
% <put your article body there>

%%%%%%%%%%%%%%%%
%% Background %%
%%
%\section*{Content}
%Text and results for this section, as per the individual journal's instructions for authors. %\cite{koon,oreg,khar,zvai,xjon,schn,pond,smith,marg,hunn,advi,koha,mouse}

\section*{1.Background}
Drug repositioning is intended to discover new uses of drugs that have been approved by drug regulatory authorities \cite{ref1}. This technology has played a major role in drug discovery because the traditional new drug development is a time-consuming, costly, and unstable process that takes 10-15 years and costs 0.8-1 billion dollars \cite{ref2,ref3,ref4}. Compared with the traditional new drug development process, the approved drugs have undergone several rigorous clinical trials, and their toxic and side effects have been strictly evaluated \cite{ref5}. Hence, drug repositioning technology can shorten the drug development cycle to 6.5 years, research and development funding could be reduced to 3 million dollars \cite{ref6,ref7}, and the related drugs can pass the regulatory review more easily\cite{ref8}.\par

The prediction of drug-target interactions is an important process in drug discovery. Targets are biological macromolecules that exert pharmacological effects in the human body and are directly related to diseases, therefore, the prediction of drug-target associations also has important research significance for drug repositioning. In recent years, many researchers have developed various computational models to predict large-scale potential drug-target associations. The research of Chen et al. \cite{ref28} not only summarized the databases and web servers involved in drug target identification and drug discovery, but also introduced some of the latest computational models for drug-target interaction prediction, which focuses on the advantages and disadvantages of network-based and machine learning based methods. Ezzat et al. \cite{ref29} introduced a chemical genomics method for calculating drug-target interaction predictions. They divided chemical genomics methods into neighborhood model based methods, local model based methods, network diffusion based methods, matrix factorization based methods and feature classification based methods. And they focused on the prediction performance of these methods in different situations. In general, it is necessary to develop novel and effective prediction methods to avoid the determination of drug-target interactions only through expensive, laborious and uncertain traditional experimental methods.\par

Recently, the graph neural network has attracted the attention of many scholars, and many researchers have applied it to the research of drug-target-disease associations. Han et al. \cite{ref30} combined graph convolutional network (GCN) and matrix factorization to propose a new disease gene association task framework GCN-MF. With the help of GCN, the framework can capture the non-linear interaction between disease and gene, and use the similarity between the measured gene and disease phenotype for prediction work. Long et al. \cite{ref31} proposed a graph convolutional network (GCN)-based framework-GCNMDA to predict human microbe-drug associations. The framework is based on a heterogeneous network of drugs and microorganisms which constructed with rich biological information. In the hidden layer of the GCN, the conditional random field (CRF) with the attention mechanism is further used to more accurately aggregate the neighborhood representations while ensuring that similar nodes (for example, microorganisms or drugs) have similar vector representations.\par

Benefited from the success of the CF (Collaborative Filtering) model in the field of recommendation systems \cite{ref9,ref10,ref11}, more and more researchers have applied the CF model to the field of drug repositioning. In general, the computational methods of drug repositioning can be categorized into two main groups. One is neighborhood based models \cite{ref12,ref13,ref14} and the other is latent factor based models \cite{ref15,ref16,ref17,ref18}.\par

Neighborhood based models recommends potential targets for drugs by identifying neighborhoods of similar drugs or diseases based on previous associations. A computational framework has been suggested by Wang et al. \cite{ref12}, HGBI, a heterogeneous drug-target graph that includes known drug-target interactions as well as similarities between drug-drug and target-target. A novel graph-based inferencing technique is implemented based on this graph to recommend potential targets to drugs. Martinez et al. \cite{ref13} created a drug-disease priority-setting methodology called DrugNet based on ProphNet, a network-based priority-setting technique. DrugNet model establishes a network of interconnected medicines, proteins and illnesses and recognizes new associations of drug-disease by disseminating data in the heterogeneous network above. Based on the theory that comparable drugs are usually associated with comparable illnesses, Luo et al. \cite{ref14} suggested a novel computational technique called MBiRW, using some extensive similarity measures and Bi-Random Walk (BiRW) to detect prospective novel signs for the specified drug.\par

Latent factor based models project each drug and disease into a common low dimensional space to capture latent associations. Gottlieb et al. \cite{ref15} suggested a model called PREDICT, which calculates the connection between future drugs and illnesses, primarily by incorporating the similarities between different drugs and illnesses and using these characteristics to acquire fresh prospective characteristics through a logical classifier. Luo et al. \cite{ref16} built a heterogeneous drug-disease interaction system by incorporating drug-drug, disease-disease, and drug-disease networks denoted with a vast adjacency matrix for drug-disease, then implement a Singular Value Thresholding algorithm to finish the adjacency matrix for drug-disease with expected results for unidentified drug-disease pairs. In order to balance the calculation error between the drug similarity and the disease similarity, Yang et al. \cite{ref17} proposed BNNR model, which incorporates the regularization of nuclear specifications into the matrix decomposition model, and can effectively solve the problem of overfitting and improve the prediction accuracy of the model. Yang et al. \cite{ref18} proposed an additional neural matrix factorization (ANMF) model, using the auxiliary information of drugs or diseases to overcome the problem of data sparsity and introducing the neural network, so that the ANMF model can capture the nonlinear relationship between drugs and diseases. \par
 
However, the above researches were based on a single type of CF model to solve the problem of drug repositioning, which can lead to the following defects. Neighborhood based methods capture local structure but usually ignore the majority of scores available owing to choosing from the junction of feedback between two drugs or diseases at most K observations. In contrast, models of latent factor capture the general global structure of the interactions between drugs and diseases, but often overlook the existence of some powerful associations. At the same time, a specific drug usually treats a smaller number of diseases to make the drug-disease correlation matrix relatively sparse. Hence relying solely on sparse data of drug-disease association can easily lead to cold start problems.\par

In recent years, due to the nonlinear fitting ability and excellent performance in mining effective hidden features from raw data, deep learning has achieved remarkable success in many fields. The memory network has achieved great achievement in the field of machine translation for its long-term and short-term memories of historical information. Hence, inspired by deep learning and the memory network \cite{ref11,ref19,ref20}, we propose the Hybrid Attentional Memory Network (HAMN), a hybrid unified model that combines the advantages of both types of CF models. At the same time, the cold-starting problem is highly challenging in the drug repositioning application scenario, which mainly refers to the lack of history data on the effects of new drugs towards other diseases. Without the historical treatment data, it is impossible to predict the corresponding treatment mechanism. So, we introduce drug-drug similarity and disease-disease similarity information to overcome cold start problems to some extent in the drug repositioning. \par

In the HAMN model, we combine the attention mechanism with memory unit \cite{ref21} to generate the neighborhood representation that captures the higher-order complex associations between drugs and diseases. Memory unit allows encoding of rich feature representations, while attention mechanisms can assign influential neighbors greater weight. Next, a variant version of the autoencoder is used to extract the valid latent factor of drug and disease and reduce the side effects of cold-starting problem by combining drug similarity, disease similarity with drug-disease associations. Finally, a nonlinear interaction between the local neighborhood representation and the global latent factors derives the predicted value.\par

Our main contributions can be summarized as follows:\par
 
(1) We propose the HAMN model, a new network framework that combines neighborhood based method with latent factor based model by the memory network, to capture both the global structural information of drug-disease associations and the local information contained in some strong drug-disease associations. \par

(2) We introduce an attention mechanism to enable influential neighbors to make greater contributions. The experimental results show that this strategy can can improve the performance of the model. \par

(3) The HAMN model has been systematically tested in two real data sets, Gottlieb dataset and Cdataset \cite{ref16}. The experimental results show that the performance of our proposed HAMN model exceeds the state-of-the-art according to the AUC, AUPR or HR indicators.\par

The rest of this paper is as constructed as follows: we will introduce the implementation details and principles of the HAMN model in section 2. In section 3, the experiments and results of the HAMN model on the Gottlieb dataset and the Cdataset will be presented, and the discussion of the experiments will be given in section 4.  The final section will serve as a summary of our work and a guideline for future ventures.\par

\section*{2.Methods}
The overall architecture of our proposed Hybrid Attentional Memory Network (HAMN) model is shown in Figure 1. At a high level, the HAMN model consists of three modules: 1) the neighborhood contribution representation module, 2) the mining latent factor module, and 3) the predictive value generation module. 

First, the neighborhood contribution representation module captures the local information contained in few strong drug-disease associations. The module derives the neighborhood contribution representation by combining the memory unit and the attention weight mechanism, which will be described in detail in section 2.1.\par

Next, the mining latent factor module captures the global information of drug-disease associations. The module uses a variant version of autoencoder to combine drug-disease relationships, drug similarity with disease similarity for the extraction of drug latent factor or disease latent factors, which will be discussed in detail in Section 2.2.\par

Finally, the predictive value generation module uses nonlinear function to calculate the predicted value by combining the latent factor of drug, the latent factor of disease and the neighborhood representation.This will be described in detail in Section 2.3. At the end of this section, we will derive the general loss function of the HAMN model and the learning of the corresponding parameters.\par

\subsection*{2.1 Neighborhood contribution representation}

In order to capture the local information contained in some strong drug-disease associations, inspired by \cite{ref20}, we first define the latent factor of drug called $drug_i $, where $drug_i \in \mathbb{R}^{1\times d} $ is generated by a set of parameter vectors, $d $ is the dimension of latent factor, which stores the characteristic information of the drug. And defined the latent factor of disease called $disease_j $, where $disease_j \in \mathbb{R}^{1\times d} $ is generated by another set of parameter vectors, which stores the specific preferences of the disease. Next we define the drug preference vector $p_{ij} $ as shown in equation (1), where each dimension $p_{ijn} $ represents the degree of similarity between the target drug $i $ and its neighbor drug $n $.\par

\begin{equation}
p_{ijn}=drug_{i}^{T}drug_n\qquad \forall n\in N\left( j \right) 
\end{equation}\par

Where $N\left( j \right)$ represents the collection of drugs that are associated with disease $j $. The intuition of our design formula (1) is as follows, the degree of compatibility between the target drug $i $ and the neighbor drug $n$ is calculated by performing the inner product operation of both the latent factor of drug $i $ and the latent factor of the neighbor drug $n$. The inner product operation enables the neighborhood drug similar to the target drug $i$ to achieve a larger compatible value, and vice versa. \par

According to the hypothesis that similar drugs can treat similar diseases, when drug $i $ infers whether it can treat the disease $j$, more similar neighbor drugs contribute more to the decisions. Hence, by formula (2) normalizing the drug preference vector $p_{ij}$, the attention weight of the target drug $q_{ij}$ can be obtained. This attention weight is used to infer the contribution weight of the neighboring drugs. It works because the attention weight vector $q_{ij}$ can impose higher weights on similar drugs in neighbors, while reducing the importance of less similar drugs, hence the target drug $i$ focuses on the influential subset of drugs in the neighborhood when making decisions.\par

\begin{equation}
q_{ijn}= \frac {\exp (p_{ijn})} {\sum_{k\in N(i)} \exp(p_{ijn})}  
\end{equation}\par

In order to learn the local information contained in a few strong drug-disease associations, inspired by the memory network and the hypothesis which the local structural information contained in the strong association is usually provided by the neighbor of the target drug, hence the HAMN model uses an external memory unit to store the characteristic information of the drug in the role of neighbor to serve as the local structural information contained in the strong drug-disease associations. Then we use the attention weight vector $q_{ij} $ to accumulate the neighborhood information contained in all the neighbor drugs of the target drug to obtain the final neighborhood contribution representation. The generation method is shown in formula (3).\par

\begin{equation}
o_{ij}=\sum_{n \in N(j)} q_{ijn}c_n 
\end{equation}\par

Where $c_n $ is another embedding vector of drug $n $, which is called external memory in the original memory network framework. The external memory allows the storage of long-term information pertaining specifically to each drug’s role in the neighborhood. Its essence is a set of parameter vectors, which can be represented by vectors $c_{n}=\left[ m_1,m_2,...m_l \right] $, where $m_l $ represents the parameters that can be learned during model training. In other words, the attention mechanism selectively weights the neighbors according to the specific drug and disease. The external memory unit $c_n $ stores the local structural information contained in the strong drug-disease associations. Then the neighborhood contribution representation generated by accumulating the sum of the attention vector and the memory unit $c_n $, which can make the contribution value of the influential neighbor greater and can capture local structural information contained in the strong drug-disease association.\par

It is worth noting that the dimension of the external memory unit does not need to be consistent with the dimension of the hidden feature vector of the drug. By adjusting the dimension of the external memory unit, it can meet different scales of computing drug repositioning data sets, which enhances the scalability of the model to a certain extent. In the experimental section 3.3.1, the effect of external memory unit dimension on model performance will be discussed.\par

\subsection*{2.2 Mining the latent factor of drugs and diseases}

Both $drug_i $ and $disease_j$ in section 2.1 are represented by parameter vectors, which required a large amount of historical drug-disease correlation data to ensure the convergence and validity of that model parameters. However, the data of computational drug repositioning is generally sparse and cannot meet the training requirements of the above parameter vectors. At the same time, the cold start problem is a major challenge in the field of computational drug repositioning. In order to extract effective latent factor and alleviate cold start problems, this section use a variant version of autoencoder to extract the latent factor of drugs and diseases instead of the above, and combine drug similarity and disease similarity.\par

The bottom of Figure 1 shows the process of mining the latent factor of drug $i$ and disease $j$. We focus on the process of mining the latent factor of drug $i$, because the process of mining the latent factor of disease $j$ is theoretically the same.\par

$R$ stands for the drug-disease associations matrix,  where $s_{i}^{drug}=\{R_{i1},R_{i2},...R_{in}\}$ represents the associations among drug $i$ and all diseases in the data set. DrugSim stands for the drugs-drugs similarity matrix, where $DrugSim_{i*}=\left[ DrugSim_{i1},DrugSim_{i2},...,DrugSim_{im} \right]$ represents the similarity between drug $i$ and $m$ drugs in the data set. To enhance the robustness of the input data, random noise is added to $s_{i}^{drug} $ and $DrugSim_{i*} $ to generate $\tilde{s}_{i}^{drug} $ and $\tilde{D}rugSim_{i*} $. Then we perform the following encoding and decoding operations on the above two inputs to extract the latent factor of the drug $i$, $drug_i $.\par

\begin{equation}
drug_i=g\left( W_1\tilde{s}_{i}^{drug}+V_1\tilde{D}rugSim_{i*}+b_d \right)   
\end{equation}\par

\begin{equation}
\hat{s}_{i}^{drug}=f\left( W_2drug_i+b_s \right)   
\end{equation}\par

\begin{equation}
\hat{D}rugSim_{i*}=f\left( V_2drug_i+b_D \right) 
\end{equation}\par

Equation (4) is the encoding operation, and equations (5) and (6) are the decoding operations, where $drug_i $ represents the latent factor of the drug $i$. $g$ and $f$ represent any activation functions, $W$ and $V$ represent weight parameters, and $b$ represents bias parameters.\par

The loss caused by the above encoding and decoding operations includes the error between all inputs and their reconstructed values, and the loss function is as shown in equation (7), where $\parallel s_{i}^{drug}-\hat{s}_{i}^{drug}\parallel ^2 $ and $\parallel DrugSim_{i*}-\hat{D}rugSim_{i*}\parallel ^2 $ represent the error caused by the input value and the reconstructed value, and $\parallel W_l\parallel ^2+\parallel V_l\parallel ^2 $ controls the complexity of the model, which improves the model’s  generalization ability. $\alpha  $ represents the equalization parameter and $\lambda$ represents the regularization parameter.\par

\begin{align}
\text{arg\;}\min_{\{W_l\},\{V_l\},\{b_l\}}\;&\alpha \parallel s_{i}^{drug}-\hat{s}_{i}^{drug}\parallel ^2 +\left( 1-\alpha \right) \parallel DrugSim_{i*}-\hat{D}rugSim_{i*}\parallel ^2 \notag \\
&+\lambda ( \sum_l{\parallel}W_l\parallel ^2+\parallel V_l\parallel ^2 )    
\end{align}\par

The latent factor of the drug $i$ can be obtained by minimizing formula (7). Similarly, the process of obtaining the latent factor of the disease $j$ is theoretically the same as the process of extracting the latent factor of the drug. The difference is that $s_{j}^{disease} $ and the diseases-diseases similarity matrix are used as inputs, where $s_{j}^{disease}=\{R_{1j},R_{2j},\cdots R_{mj}\} $ represents the vector of relationships among the disease j and all drugs in the data set.\par

\subsection*{2.3 Predictive value generation}

As mentioned above, the neighborhood based model captures the information contained in few strong drug-disease associations and the latent factor model captures the global structural information of drug-disease associations. Therefore, we used $o_{ij}$ to capture the local information of the drugs-diseases relationships, and used $drug_i$ and $disease_j $ to capture the global information of the drugs-diseases relationships, which are finally nonlinearly integrated by using the following formula (8).\par

\begin{equation}
\hat{r}_{ij}=F_{out}\left(\eta h^T\left( drug_i\odot disease_j \right) + (1-\eta)W^To_{ij}+b \right)  
\end{equation}\par

$drug_i $ and $disease_j $ represent the latent factors of drugs and diseases calculated by the HAMN model, $\odot$ represents elementwise product, and $o_{ij} $ represents neighbor contribution representation. $h$ and $W$ represent the weight parameters, $\eta $ is the balance parameter, which controls the weight of the latent factor model and the neighbor model in the final output. $b$ represents the offset parameter, $F_{out}$ represents any activation function, and $\hat{r}_{ij}$ represents the predicted value.\par

Where $h^T\left( drug_i\odot disease_j \right)$ represents the output value of the latent factor model, $W^To_{ij} $ represents the output value of the neighbor model, and equation (8) smooths the nonlinear integration of the two to obtain the predicted value. The above operation enables the HAMN model to capture both global and local information.\par

\subsection*{2.4 Parameter learning}

In this part, we will derive the final loss function of the HAMN model and the learning process of the corresponding parameters. In general, the loss function of the HAMN model includes the loss of the extracted drug and the disease latent factor and the loss between the predicted value and the target value.\par

The loss function of the extracted drug and disease latent factor are shown in equations (9) and (10), which has been derived in 2.2.\par

\begin{align}
\mathcal{L}_d&=\sum_{i}\alpha \parallel s_{i}^{drug}-\hat{s}_{i}^{drug}\parallel ^2 \notag \\
&+\left( 1-\alpha \right) \parallel DrugSim_{i*}-\hat{D}rugSim_{i*}\parallel ^2  \notag \\
&+\lambda ( \sum_l{\parallel}W_l\parallel ^2+\parallel V_l\parallel ^2 ) 
\end{align}\par

\begin{align}
\mathcal{L}_p&=\sum_{j}\beta \parallel s_{j}^{disease}-\hat{s}_{j}^{disease}\parallel ^2 \notag \\
&+\left( 1-\beta \right) \parallel DiseaseSim_{j*}-\hat{D}iseaseSim_{j*}\parallel ^2  \notag \\
&+\delta ( \sum_d{\parallel}W_d\parallel ^2+\parallel V_d\parallel ^2 )  
\end{align}\par

The loss between the predicted value and the target value is as shown in equation (11), where $r_{ij}$ represents the target value, $\hat{r}_{ij}$ is the predicted value derived from the HAMN model. In addition, $R^+$ represents the positive sample set in which from known drug-disease associations. $R^-$ represents the negative sample set, which can be obtained using negative sampling techniques \cite{ref22}.\par

\begin{equation}
\mathcal{L}_{r}=\sum_{\left( i,j \right) \in R^+\cup R^-} r_{ij}\log \hat{r}_{ij}+\left( 1-r_{ij} \right) \log \left( 1-\hat{r}_{ij} \right) 
\end{equation}\par

Hence, the final loss function of the HAMN model is shown in equation (12), where 

\begin{align}
\mathcal{L} =\mathcal{L}_{r} + \varphi \mathcal{L}_d +\psi \mathcal{L}_p  
\end{align}\par

The relevant parameters of the HAMN network can be learned by minimizing the formula (12) by the SGD algorithm.\par

As we can see from the above analysis, the model we propose has the following advantages. First, at the section 2.1, the introduction of attention weight mechanism enables the model to impose higher weight on similar drugs in neighbors, ensuring it makes a greater contribution in the decision-making stage. Finally, the linear function is used to integrate the latent factor and the neighborhood representation, so that the model has a holistic view of the drugs-diseases interactions to infer the predicted value.\par

\section*{3. Results}

This section systematically evaluates the performance of the HAMN model on two real data sets and the experimental comparisons with the most advanced algorithms currently relevant. First, the two real data sets used in the experiment will be introduced in detail in Section 3.1. Next, the evaluation criteria and calculation methods used in the experiment will be introduced in Section 3.2. Then in section 3.3, we discuss the details and specific setting values of all hyperparameters in the HAMN model, as well as the experimental analysis and the discussion of two important parameters. At the same time, in order to verify the effectiveness and superiority of the HAMN model, the HAMN model is experimentally compared with several currently relevant most advanced algorithms in section 3.4, and a detailed ablation study is also given in section 3.4. To further verify the practicability of the HAMN model, its performance in new drug scenarios will be evaluated in Section 3.5.\par

\subsection*{3.1 Data set}

This experiment uses two mainstream data sets, Gottlieb dateset and Cdataset \cite{ref16}. Gottlieb dateset contains 593 drugs, 313 diseases and 1933 proven drug-disease relationships. Cdataset contains 663 drugs, 409 diseases and 2,532 proven drug-disease relationships. See Table 1 and 2 for details. The drugs and diseases contained in the above data sets were registered in DrugBank \cite{ref23} and Online Mendelian Inheritance in Man \cite{ref24} respectively.\par

\begin{table}[h!]
	\caption{Statistics of the Gottlieb dataset.}
	\begin{tabular}{ccccc}
		\hline
		Dataset& Drugs& Diseases & Interactions & Sparsity\\ \hline
		Gottlieb & 593 & 313 & 1933 & $1.041\times10^{-2}$\\ \hline
	\end{tabular}
	\label{table1}
\end{table}\par

\begin{table}[h!]
	\caption{Statistics of the Cdataset.}
	\begin{tabular}{ccccc}
		\hline
		Dataset& Drugs& Diseases & Interactions & Sparsity\\ \hline
		Cdataset & 663 & 409 & 2532 & $9.337\times10^{-3}$\\ \hline
	\end{tabular}
	\label{table2}
\end{table}\par

Drug similarities are calculated on the basis of SMILES \cite{ref25} using the Chemical Development Kit \cite{ref26}. Pairwise drug resemblance and chemical structures are referred to as their 2D chemical patterns Tanimoto score. MimMiner \cite{ref27}, which estimates the degree of pairwise disease resemblance through text mining their medical description data in the OMIM database, obtains the similarities among illnesses. In addition, both drug-drug similarity and disease-disease similarity take into account the prior relationship between drugs and disease.\par

\subsection*{3.2 Evaluation metrics}

This experiment uses a ten-fold cross-validation technique. And the unverified drug-disease relationships in the data set were taken as negative samples and placed in the test set, and then the training set is used to learn the relevant parameters of the model. The performance of the trained model on the test set is evaluated, thereby achieving a 10-fold cross-validation and final performance evaluation.\par

In order to comprehensively evaluate the performance of the HAMN model, we use AUC (Area Under Curve Area), AUPR (Area Under Precision-Recall Curve) and HR (Hit Ratio) as the evaluation indicators. AUC is currently a mainstream evaluation indicator, but for the category imbalance problem, the AUC indicator cannot capture all the information of the model, and the true performance of the model can be reflected in a more comprehensive way by adding the AUPR indicator. At the same time, HR is the most popular evaluation indicator in the field of recommendation systems, which can well reflect the performance of the model in real demand scenarios. Combined with the above three evaluation indicators, the performance of the HAMN model can be more fairly and comprehensively displayed.\par

\subsection*{3.3 Parameter settings}
The two important parameters of the HAMN model are the dimension of the memory unit $c_n$ and the balance parameter $\eta $ . Since the memory cell $c_n$ vector stores the characteristic information of the drug in the neighbor role, its size controls the complexity and fitting ability of the neighborhood module of the HAMN model. At the same time, the hyperparameters $\eta $  balance the weight ratio of the latent factor model and the neighborhood model in the final output. Appropriate values can improve the performance of the model. Therefore, this section sets up two related experiments to evaluate the performance of the HAMN model under different dimensions of the memory cell vector $c_n$ and hyperparameters $\eta $ .\par

All hyperparameters of the HAMN model are set based on their performance on the validation set. The validation set is created based on \cite{ref18}. For the dimension of memory unit and the value of $\eta $, we use the grid search to find the optimal combination in the interval $\left\{16, 32, 64, 128, 256\right\}$ and the interval $\left\{0.1,0.3,0.5,0.7,0.9\right\}$. Similarly, $\alpha$ and $\beta$ are all grid searched in the interval $\left\{ 0.1,0.3,0.5,0.7,0.9\right\}$. Besides, $\lambda$ and $\delta$ are all grid searched in the interval $\left\{0.1, 0.01, 0.001\right\}$. \par

\subsubsection*{3.3.1 The dimension of external memory unit}
The dimension of the memory unit $c_n $ is one of the important parameters of the HAMN model, which controls the complexity of the neighborhood module of the HAMN model and its learning ability. If the dimension setting is too large, the model training time will increase exponentially and over-fitting will easily occur. Conversely, setting the dimension too small will prevent the model from learning the structural information contained in some strong drug-disease associations, which will affect the performance of the neighborhood module. Therefore, this experiment is set up to observe the effect of different memory cell vector dimensions on the performance of the HAMN model. In addition, the search interval of the dimension of the memory unit is set to $\left\{16, 32, 64, 128, 256\right\}$, and the remaining hyperparameters are set to the default values. The experimental data set uses the Gottlieb data set, the evaluation index uses the AUC value.\par

Figure 2(a) shows the impact of different memory unit dimensions on the performance of the HAMN model. The abscissa of the graph represents the dimensions of the memory unit and the ordinate is the AUC value. The experimental results show that the performance of the model improves steadily with the increase of the dimension of the memory unit. When the dimension is 64, the performance of the model reaches its peak. However, followed by a degradation potentially due to overfitting and the model's AUC value begins to decrease.\par

By analyzing the above experimental results, it can be concluded that the appropriate memory unit dimension can enhance the fitting ability of the HAMN model neighborhood module, and learn the structural information of strong drug-disease correlation, thereby further improving the overall performance of the HAMN model.\par

\subsubsection*{3.3.2 The weight value of $\eta $ }

The hyperparameter $\eta$ controls the weight ratio of the latent factor module and the neighborhood module in the final output. Appropriate values are crucial to the performance of the HAMN model. Therefore, the following experiments are set up to observe the effect of different values on the performance of the HAMN model. In addition, the search interval of $\eta $ values is set to $\left\{0.1,0.3,0.5,0.7,0.9\right\}$, and the remaining hyperparameters are set to the default values. The experimental data set and evaluation indicators are consistent with Section 3.3.1.  \par

The experimental results in Figure 2(b) show that as the value of the hyperparameter $\eta$ increases continuously, the performance of the HAMN model behaves a stable linear improvement. The above experimental results show that the importance of the hidden feature module is higher than that of the neighborhood module, and it should be given higher weight. However, the neighborhood model can accurately judge part of the test set samples and the hidden feature module cannot accurately predict the part of the samples. Hence, the neighborhood model should be given partial weights so that the final predicted value takes into account the contribution of the neighborhood module. Therefore, when the value is set to 0.7, the prediction effect and generalization performance of the HAMN model are improved to a certain extent.\par

\subsection*{3.4 Method comparison}
We compare the HAMN model with several current mainstream algorithms, including the latent factor based methods and the neighborhood based methods. \par

* ANMF \cite{ref18}: The ANMF model is a neural matrix decomposition model, which is a HAMN model without neighborhood information essentially.\par

* BNNR \cite{ref17}: The BNNR model is one of the latest research achievements in the field of computational drug relocation, and its essence is a model based on hidden features. In order to balance the calculation error between the similarity between drugs and the similarity between diseases, it incorporates the regularization of nuclear specifications into the matrix decomposition model, which can effectively solve the problem of overfitting and improve the prediction accuracy of the model.\par

* DRRS \cite{ref16}: The DRRS model is a mainstream latent factor model, which uses drug-disease relationships matrix, drug similarity matrix and disease similarity matrix to generate a hybrid matrix, and then uses the SVT algorithm to matrix decompose to generate predicted values.\par

* HGBI \cite{ref12}: HGBI is a classic neighborhood based method. HGBI is introduced based on the guilt-by-association principle, as an intuitive interpretation of information flow on the heterogeneous graph.\par

The parameters of the above comparison methods are provided by their corresponding documents.\par

Table 3 and Table 4 is the experimental results of the above model on the two published data sets. No matter indicators we use, AUC, AUPR or HR metric, the HAMN model we propose outperforms other comparison methods. In terms of AUC value, the HAMN model achieved the highest value of 0.946 on the Gottlieb dataset, which was higher than the 0.938 in ANMF model, 0.932 in BNNR, 0.93 in the DRRS model and 0.829 in the HGBI model. The HAMN model also gets the highest value of 0.958 on the Cdataset.\par

In terms of AUPR value, the HAMN model achieved the highest value of 0.385 on the Gottlieb dataset, which was higher than 0.347 in the ANMF model, 0.315 in the BNNR model, 0.292 in the DRRS model and 0.16 in the HGBI model. The HAMN model also gets the highest value of 0.426 on the Cdataset.\par

In terms of HR value, the HAMN model achieved the highest value on both Gottlieb dataset and Cdataset. In the HR@10 scenario, the HAMN model achieved the highest value of 76.2\% in the Gottlieb dataset, which was higher than 74.2\% of the ANMF model, 75.9\% of the BNNR model, 72.7\% of the DRRS model and 59.3\% of the HGBI model. The HAMN model also gets the highest value of 79.1\% on the Cdataset.\par

\begin{table}[h!]
	\caption{Prediction results of different methods on Gottlieb dataset.}
	\begin{tabular}{cccccc}
		\hline
	    Method Name & AUC            & AUPR           & HR@1            & HR@5            &  HR@10                 \\ \hline
		HAMN        & \textbf{0.946} & \textbf{0.385} & \textbf{51.5\%} & \textbf{66\%}   &  \textbf{76.3\%}       \\ \hline
		ANMF        & 0.938 		 & 0.347 		  & 47.9\%			& 61.3\% 		  &  74.2\%       		   \\ \hline
		BNNR        & 0.932          & 0.315          & 50.2\%          & 64.7\%          &  75.9\%                \\ \hline
		DRRS        & 0.93           & 0.292          & 45.9\%          & 53.1\%          &  72.7\%                \\ \hline
		HGBI        & 0.829          & 0.16           & 33\%          	& 45.4\%          &  59.3\%                \\ \hline
	\end{tabular}
	\label{table3}
\end{table}\par

\begin{table}[h!]
	\caption{Prediction results of different methods on Cdataset.}
	\begin{tabular}{cccccc}
		\hline
		Method Name & AUC            & AUPR           & HR@1              & HR@5            &  HR@10                 \\ \hline
		HAMN        & \textbf{0.958} & \textbf{0.426} & \textbf{43.8\%}   & \textbf{67.2\%} &  \textbf{79.1\%}       \\ \hline
		ANMF        & 0.952 		 & 0.394 		  & 42.1\%   		  & 65.1\% 			&  76.3\%       		 \\ \hline
		BNNR        & 0.948          & 0.388          & 42.9\%            & 66.1\%          &  78.2\%                \\ \hline
		DRRS        & 0.947          & 0.351          & 32.3\%            & 59\%            &  70.1\%                \\ \hline
		HGBI        & 0.858          & 0.204          & 26.7\%            & 37.1\%          &  55.1\%                \\ \hline
	\end{tabular}
	\label{table5}
\end{table}\par 

According to the above experimental results, the HAMN model performs better than the neighborhood based model HGBI and the latent factor based model BNNR, DRRS and ANMF, which reveals the effectiveness of combining the two CF models into a single hybrid model. It is worth noting that the HAMN model is superior to the ANMF model, the latter is essentially a HAMN model without the memory unit. It reveals that the integration of neighborhood information improves the performance of the HAMN model to a certain extent. \par

\subsection*{3.5 The new drug scenario}

The new drug scenario describes the situation of predicting potential target for drug without previously known disease associations, this is more in line with the real world needs. There are 171 drugs in the Gottlieb dataset associated with only one known disease, and 177 drugs in the Cdataset associated with only one known disease. We removed drugs with one known association from the data set and placed it in the test set. The remaining drug-disease associations were used as training sets, and the model was trained and tested according to the above. The experimental parameters are set according to the rules in Section 3.3. \par

Table 5 and Table 6 list the experimental results of the above models for new drugs on the Gottlieb dataset and Cdataset. No matter indicators we use, AUC, AUPR or HR metric, the HAMN model we propose performs better than other comparison methods. In terms of AUC value, the HAMN model achieved the highest value of 0.881 on the Gottlieb dataset, which was higher than the 0.859 in ANMF model, 0.83 in the BNNR model, 0.824 in the DRRS model and 0.746 in the HGBI model. The HAMN model also gets the highest value of 0.869 on the Cdataset.\par

\begin{table}[h!]
	\caption{Prediction results of different methods for new drug on Gottlieb dataset.}
	\begin{tabular}{cccccc}
		\hline
		Method Name & AUC            & AUPR           & HR@1            & HR@5            &  HR@10                 \\ \hline
		HAMN        & \textbf{0.881} & \textbf{0.193} & \textbf{30.4\%} & \textbf{36.8\%} &  \textbf{49.7\%}       \\ \hline
		ANMF        & 0.859 		 & 0.161		  & 28.1\%   		& 34.5\% 		  &  46.2\%      \\ \hline
		BNNR        & 0.83           & 0.142          & 28.7\%          & 35.1\%          &  47.4\%                  \\ \hline
		DRRS        & 0.824          & 0.107          & 28.1\%          & 30.4\%          &  39.2\%                \\ \hline
		HGBI        & 0.746          & 0.065          & 9\%          	& 14\%            &  24.6\%                \\ \hline
	\end{tabular}
	\label{table4}
\end{table}\par

\begin{table}[h!]
	\caption{ Prediction results of different methods for new drug on Cdataset.}
	\begin{tabular}{cccccc}
		\hline
		Method Name & AUC            & AUPR           & HR@1              & HR@5              &  HR@10                 \\ \hline
		HAMN        & \textbf{0.869} & \textbf{0.113} & 26\%  		  	  & \textbf{35\%}     &  \textbf{39.5\%}       \\ \hline
		ANMF        & 0.857 		 & 0.097 		  & 19.2\%  		  & 33.3\%   		  &37.3\%      \\ \hline
		BNNR        & 0.837          & 0.091          & 25.4\%            & 33.9\%            &  38.4\%                  \\ \hline
		DRRS        & 0.824          & 0.084          & 25.4\%   		  & 30.5\%            &  35\%                  \\ \hline
		HGBI        & 0.732          & 0.022          & 11.3\%            & 21.5\%            &  26\%                  \\ \hline
	\end{tabular}
	\label{table6}
\end{table}\par

In terms of AUPR values, the HAMN model achieved the highest value of 0.193 on the Gottlieb dataset, which was higher than 0.161 for the ANMF model, 0.142 for the BNNR, 0.107 for the DRRS model and 0.065 for the HGBI model. In addition, the HAMN model gets the highest value of 0.113 on the Cdataset.\par

In terms of HR value, the HAMN's value of HR@1 is smaller than the DRRS model on Cdataset. The possible reason is the sparseness of the data set. However, in the case of HR@5, HR@10, the HAMN model has achieved the highest value. In the HR@10 scenario, the HAMN model achieved the highest value of 49.1\% on the Gottlieb dataset, which was higher than the 46.2\% in ANMF model, 47.4\% in the BNNR model, 39.2\% in the DRRS model and 24.6\% of the HGBI model. Moreover, HAMN model achieves the maximum value of 39.5\% on Cdataset.\par

Given the inherent nature of sparse data and cold start problems, new drug scenario has always been a major difficulty in computing drug relocation. Moreover, the new drug scene is more in line with the needs of the real world, researchers are more and more incentivized to solve this problem. Different from the previous models that only use sparse historical drug-disease association, the HAMN model also introduces similarity between drugs, similarity between diseases and structural information contained in some strong correlations, which can alleviate the cold start problem. The above experimental results demonstrated that the proposed HAMN model can alleviate the cold start problem to some extent due to the inclusion of auxiliary information and neighbor information. Therefore, the HAMN model can be applied to new drug scenarios. \par

\section*{4.Discussion}

As seen in the experimental results on two mainstream data sets in the real world, the HAMN model has outperformed the most advanced algorithms in terms of the indicators AUC, AUPR and HR. For the Gottlieb data set, the AUC, AUPR, and HR values were 0.946, 0.385, and 76.2\% respectively. The prediction performance of the model for Cdataset is 0.958, the AUPR value is 0.426, and the HR value is 79.1\%. The validity and superiority of the HAMN model are verified to some extent by the fact that the above results are better than the comparison mainstream algorithms.\par

Finally comparing with the ANMF model, essentially a HAMN model without neighborhood information, HAMN Model can improve the performance of the algorithm to a certain extent and outperformed the ANMF model in all evaluation index. \par

\section*{5.Conclusion}
Computational drug repositioning, which aims to find new applications for existing drugs, is gaining more attention from the pharmaceutical companies due to its low attrition rate, reduced cost, and shorter timelines for novel drug discovery. In this work, we developed a novel network architecture HAMN for drug repositioning. HAMN model uses a memory network to combine the neighborhood based approaches with latent factor based models in a nonlinear manner, and incorporates drug-disease auxiliary information to alleviate the cold start problem. Experimental results on two data sets demonstrated that the HAMN model we proposed outperformed the other state of art methods. In future works, we will delve into the use of multi-source data to calculate the similarity between drugs and diseases and more types of latent factor models or neighborhood based approaches to further improve the performance of the model.\par

%%%%%%%%%%%%%%%%%%%%%%%%%%%%%%%%%%%%%%%%%%%%%%
%%                                          %%
%% Backmatter begins here                   %%
%%                                          %%
%%%%%%%%%%%%%%%%%%%%%%%%%%%%%%%%%%%%%%%%%%%%%%

\begin{backmatter}

%%  After submission of the .TEX file,                     %%
%%  you will be prompted to submit your .BBL file.         %%
\section*{Declarations}

\section*{Abbreviations}
CF: Collaborative Filtering;
HAMN: Hybrid Attentional Memory Network;
GCN: graph convolutional network;
CRF: conditional random field;
ADAE: Additional stacked denoising autoencoder; ANMF: Additional neural
matrix factorization; AUC: Area under curve; AUPR: Area under precision-recall
curve; CDK: Chemical development kit; DRRS: Drug repositioning
recommendation system; FDA: The US food and drug administration; FN: False
negative; FP: False positive; FPR: False positive rate; GMF: Generalized matrix
factorization; HGBI: Heterogeneous graph based inference; HR: Hit ratio; HR@n:
Hit ratio with cut offs at n; NMF: Non-negative matrix factorization; OMIM:
Online mendelian inheritance in man; ROC: Receiver operating characteristic;
SGD: Stochastic gradient descent method; SMILES: Simplified molecular input
line entry specification; SVT: Fast singular value thresholding algorithm; TN:
True negative; TP: True positive; TPR: True positive rate; 10-CV: Ten-fold cross
validation.

\section*{Ethics approval and consent to participate}
Not applicable.

\section*{Competing interests}
The authors declare that they have no competing interests.

\section*{Author's contributions}
HJY contributed to the design of the study. YXX designed and implemented the HAMN method, performed the experiments, and drafted the manuscript. ZI and GZ contributed to improving the writing of manuscripts. All authors read and approved the final manuscript.

\section*{Acknowledgements}
We would like to express our deepest gratitude and appreciation for all reviewers and editors. 

\section*{Funding}
This work has been supported by the National Key R\&D Program of China(2019YFC1711000) and Collaborative Innovation Center of Novel Software Technology and Industrialization. 

\section*{Availability of data and materialss}
The datasets that support the findings of this study are available in https://github.com//bioinfomaticsCSU/MBiRW.

\section*{Consent for publication}
Not applicable.

\begin{figure}[h]
	\caption{\csentence{The architecture of the HAMN model.}}
	\includegraphics[scale=0.9]{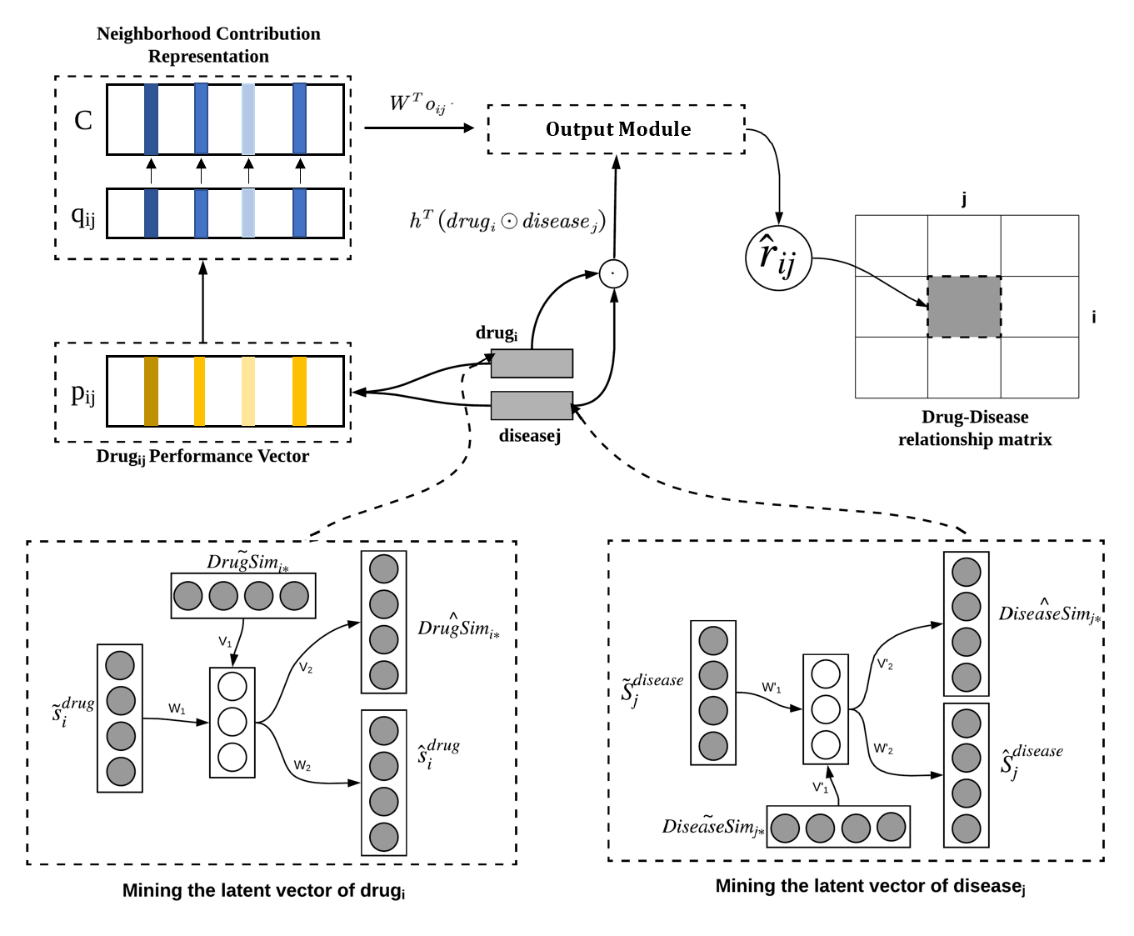}
	\label{figure1}
\end{figure}\par

\begin{figure} [h!]
	\centering 
	\subfigure[]
	{ 
		%\label{fig:subfig:a} %% label for first subfigure 
		\includegraphics[scale=0.3]{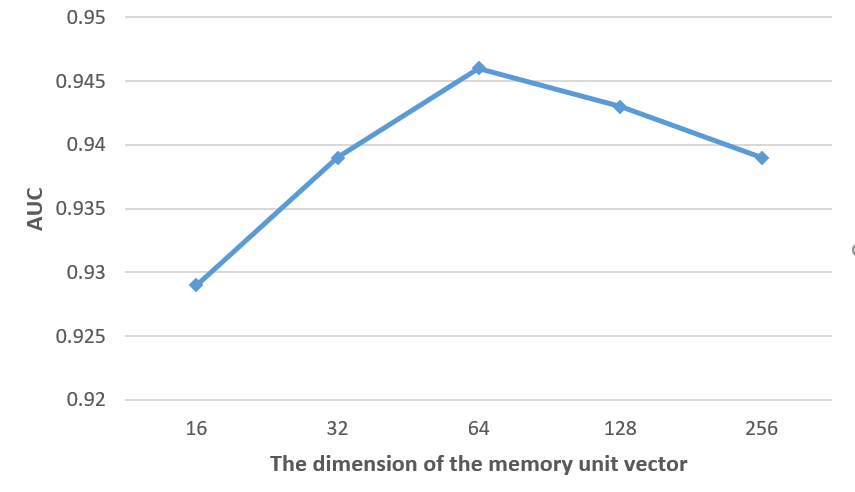} 
	} 
	\subfigure[]
	{ 
		%\label{fig:subfig:b} %% label for second subfigure 
		\includegraphics[scale=0.3]{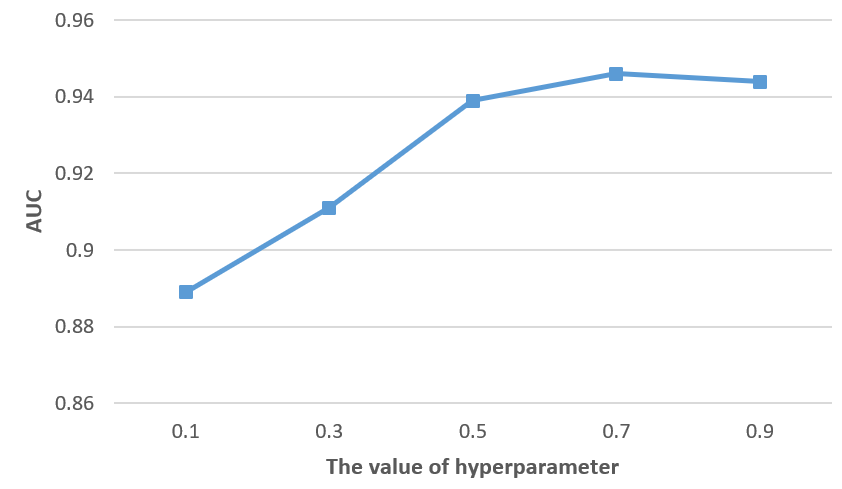} 
	} 
	\caption{\small(a)The effect of the dimensions of the external memory unit vector on the HAMN model. (b)The effect of hyperparameters $\eta $ on HAMN model.}
	\label{fig2}
	%\label{fig:subfig} %% label for entire figure 
\end{figure}\par
% if your bibliography is in bibtex format, use those commands:

% for author-year bibliography (bmc-mathphys or spbasic)
% a) write to bib file (bmc-mathphys only)
% @settings{label, options="nameyear"}

% or include bibliography directly:
% \begin{thebibliography}
% \bibitem{b1}
% \end{thebibliography}

%%%%%%%%%%%%%%%%%%%%%%%%%%%%%%%%%%%
%%                               %%
%% Figures                       %%
%%                               %%
%% NB: this is for captions and  %%
%% Titles. All graphics must be  %%
%% submitted separately and NOT  %%
%% included in the Tex document  %%
%%                               %%
%%%%%%%%%%%%%%%%%%%%%%%%%%%%%%%%%%%

%%
%% Do not use \listoffigures as most will included as separate files

\end{backmatter}
\end{document}